\documentclass[12pt,a4paper]{scrartcl}

\DeclareOldFontCommand{\bf}{\normalfont\bfseries}{\mathbf}

\usepackage[utf8]{inputenc}

\usepackage{cite}

\addtokomafont{disposition}{\rmfamily}
\usepackage{newpxtext,newpxmath}

\addtokomafont{disposition}{\rmfamily}

\usepackage{nameref,hyperref}

\usepackage[right]{lineno}

\usepackage{microtype}

\usepackage{changepage}

\usepackage[aboveskip=1pt,labelfont=bf,labelsep=period,singlelinecheck=off]{caption}

\makeatletter
\renewcommand{\@biblabel}[1]{\quad#1.}
\makeatother

\usepackage{color}

\definecolor{Gray}{gray}{.25}

\usepackage{graphicx}

\usepackage{sidecap}

\usepackage{wrapfig}
\usepackage[pscoord]{eso-pic}
\usepackage[fulladjust]{marginnote}
\reversemarginpar

\usepackage{url,hyperref,lineno,microtype,subcaption}
\usepackage{textcomp}
\usepackage[onehalfspacing]{setspace}
\usepackage[english]{babel}
\usepackage[capitalize, noabbrev]{cleveref}
\usepackage{natbib}

\begin{document}

\begin{flushleft}
{\Large
\textbf\newline{Tracking all members of a honey bee colony over their lifetime using learned models of correspondence}
}
\newline
\\
Franziska Boenisch\textsuperscript{1},
Benjamin Rosemann\textsuperscript{1},
Benjamin Wild\textsuperscript{1},
Fernando Wario\textsuperscript{1},
David Dormagen\textsuperscript{1},
Tim Landgraf\textsuperscript{1,*},
\\
\bigskip
\bf{1} Dahlem Center of Machine Learning and Robotics, Dept. Mathematics and Computer Science, Freie Universit\"at Berlin, Berlin, Germany 
\\
\bigskip
* tim.landgraf@fu-berlin.de

\end{flushleft}

\section*{Abstract}
Computational approaches to the analysis of collective behavior in social insects increasingly rely on motion paths as an intermediate data layer from which one can infer individual behaviors or social interactions. Honey bees are a popular model for learning and memory. Previous experience has been shown to affect and modulate future social interactions. So far, no lifetime history observations have been reported for all bees of a colony. In a previous work we introduced a recording setup customized to track up to 4000 marked bees over several weeks. Due to detection and decoding errors of the bee markers, linking the correct correspondences through time is non-trivial. In this contribution we present an in-depth description of the underlying multi-step algorithm which produces motion paths, and also improves the marker decoding accuracy significantly. The proposed solution employs two classifiers to predict the correspondence of two consecutive detections in the first step, and two tracklets in the second. We automatically tracked \texttildelow 2000 marked honey bees over 10 weeks with inexpensive recording hardware using markers without any error correction bits. We found that the proposed two-step tracking reduced incorrect ID decodings from initially \texttildelow 13\% to around 2\% post-tracking. Alongside this paper, we publish the first trajectory dataset for all bees in a colony, extracted from \texttildelow 3 million images covering three days. We invite researchers to join the collective scientific effort to investigate this intriguing animal system. All components of our system are open-source.

\section{Introduction}

Social insect colonies are popular model organisms for self-organization and collective decision making. Devoid of central control, it often appears miraculous how orderly termites build their nests or ant colonies organize their labor. Honey bees are a particularly popular example - they stand out due to a rich repertoire of communication behaviors \citep{vonfrisch1965,seeley2010} and their highly flexible division of labor \citep{robinson1992,johnson2010}. A honey bee colony robustly adapts to changing conditions, whether it may be a hole in the hive that needs to be repaired, intruders that need to be fended off, brood that needs to be reared, or food that needs to be found and processed. The colony behavior emerges from interactions of many thousand individuals. The complexity that results from the vast number of individuals is increased by the fact that bees are excellent learners: empirical evidence indicates that personal experience can modulate communication behavior \citep{gruter2006,balbuena2012,gruter2009,gruter2011,goyret2005,de-marco2001,richter1993}. Especially among foragers, personal experience may be very variable. The various locations a forager visits might be dispersed over large distances (up to several kilometers around the hive) and each site might offer different qualities of food, or even pose threats. Thus, no two individuals share the same history and experiences. Evaluating how personal experience shapes the emergence of collective behavior and how individual information is communicated to and processed by the colony requires robust identification of individual bees over long time periods.

However, insects are particularly hard to distinguish by a human observer. Tracking a bee manually is therefore difficult to realize without marking these animals individually. Furthermore, following more than one individual simultaneously is almost impossible for the human eye. Thus, the video recording must be watched once per individual, which, in the case of a bee hive, might be several hundred or thousand times. Processing long time spans or the observation of many bees is therefore highly infeasible, or is limited to only a small group of animals. Most studies furthermore focused on one focal property, such as certain behaviors or the position of the animal.
Over the last decades, various aspects of the social interactions in honey bee colonies have been investigated with remarkable efforts in data collection: Naug \citep{naug2008} manually followed around 1000 marked bees in a one hour long video to analyze food exchange interactions. Barachi and Cini \citep{baracchi2014} manually extracted the positions of 211 bees once per minute for 10 hours of video data to analyze the colony\textquotesingle s proximity network. Biesmeijer and Seeley \citep{biesmeijer_use_2005} observed foraging related behaviours of a total of 120 marked bees over 20 days. Couvillon and coworkers manually decoded over 5000 waggle dances from video \citep{Couvillon2014waggle}. Research questions requiring multiple properties, many individuals, or long time frames are limited by the costs of manual labor. 

In recent years, computer vision software for the automatic identification and tracking of animals has evolved into a popular tool for quantifying behavior \citep{krause2013,dell2014}.
Although some focal behaviors might be extracted from the video feed directly \citep{berman2014mapping,wiltschko2015mapping,wario2017automatic}, tracking the position of an animal often suffices to infer its behavioral state \citep{kabra2013jaaba, eyjolfsdottir2016learning, blut_automated_2017}.
Tracking bees within a colony is a particularly challenging task due to dense populations, similar target appearance, frequent occlusions and a significant portion of the colony frequently leaving the hive. The exploration flights of foragers might take several hours, guard bees might stay outside the entire day to inspect incoming individuals. The observation of individual activity over many weeks, hence, requires robust means for unique identification.

For a system that robustly decodes the identity of a given detection, the tracking task reduces to simply connecting matching IDs. Recently, three marker-based insect tracking systems \citep{Mersch2013, CrallBEEtagLowCostImageBased2015, gernat_automated_2018} have been proposed that use a binary code with up to 26 bits for error correction \citep{thompson1983error}. The decoding process can reliably detect and correct errors, or, reject a detection that can not be decoded. There are two disadvantages to this approach. First, error correction requires relatively expensive recording equipment (most systems use at least a 20 MP sensor with a high quality lens). Second, detections that could not be decoded can usually not be integrated into the trajectory, effectively reducing the detection accuracy and sample rate.

\begin{figure}[h!]
\begin{center}
    \includegraphics[width=15cm]{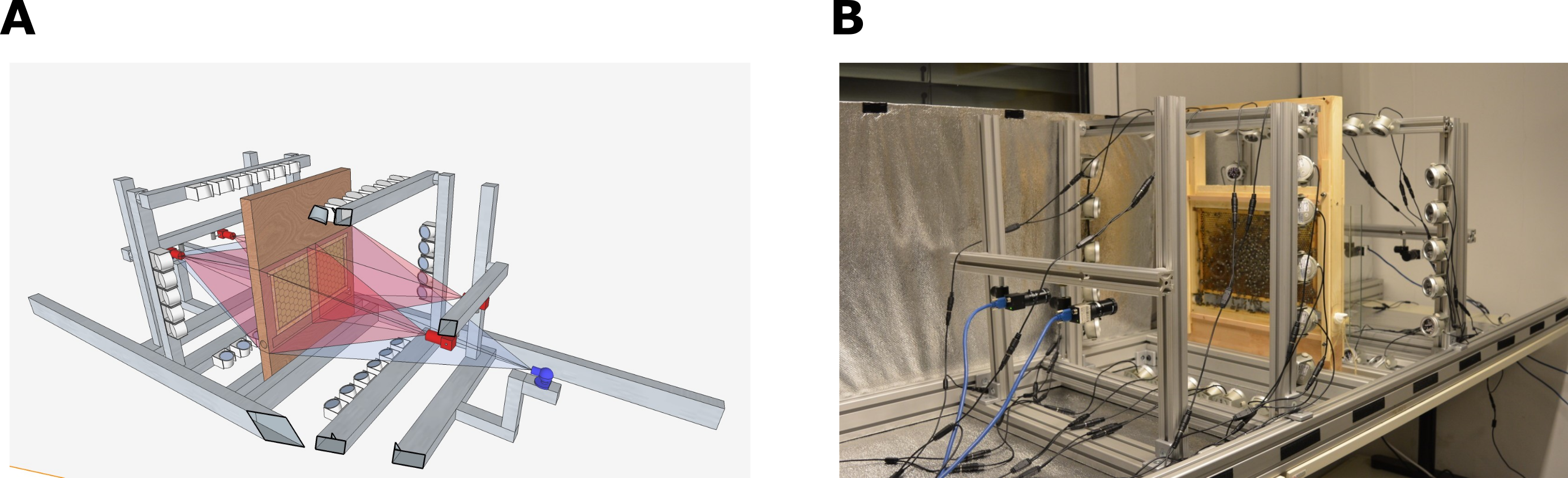}
\end{center}
\caption{(\textbf{A}) Schematic representation of the setup. Each side of the comb is recorded by two 12 MP PointGrey Flea3 cameras. The pictures have an overlap of several centimeters on each side. (\textbf{B}) The recording-setup used in summer 2015. The comb, cameras and the infrared lights are depicted, the tube that can be used by the bees to leave the setup is not visible. During recording, the setup is covered. Figures adapted from \citep{wario_automatic_2015}.}\label{fig:setup}
\end{figure}

\begin{figure}[h!]
\begin{center}
    \includegraphics[width=10cm]{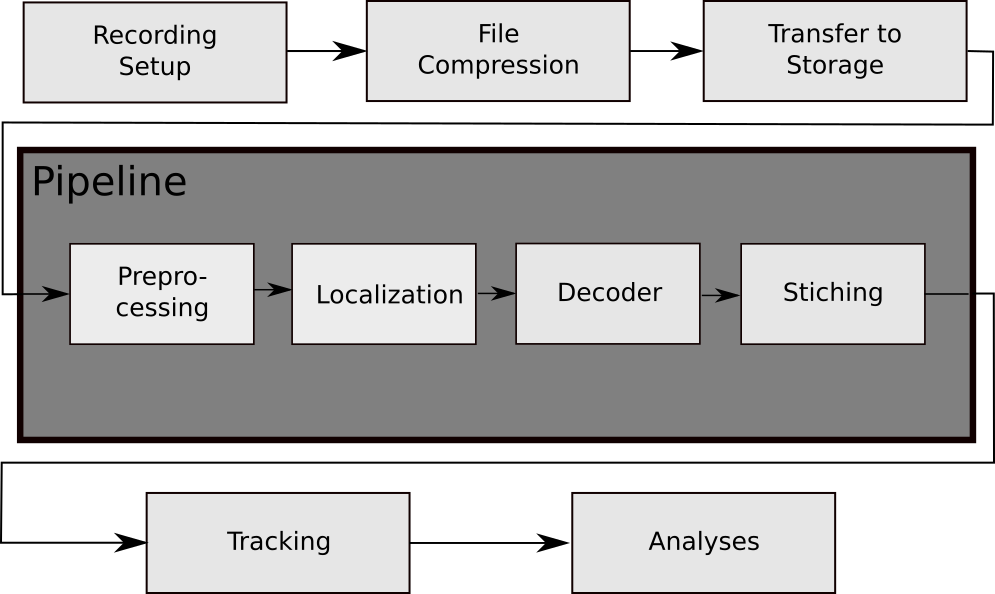}
\end{center}
\caption{ The data processing steps of the BeesBook project. The images captured by the recording setup are compressed on-the-fly to videos containing 1024 frames each. The video data is then transferred to a large storage from where it  can be accessed by the pipeline for processing. Preprocessing: histogram equalization and subsampling for the localizer. Localization: bee markers are localized using a convolutional neural network. Decoding: a second network decodes the IDs and rotation angles. Stitching: the image coordinates of the tags are transformed to hive coordinates and duplicate data in regions where images overlap are removed.}\label{fig:pipeline}
\end{figure}

\begin{figure}[h!]
\begin{center}
    \includegraphics[width=15cm]{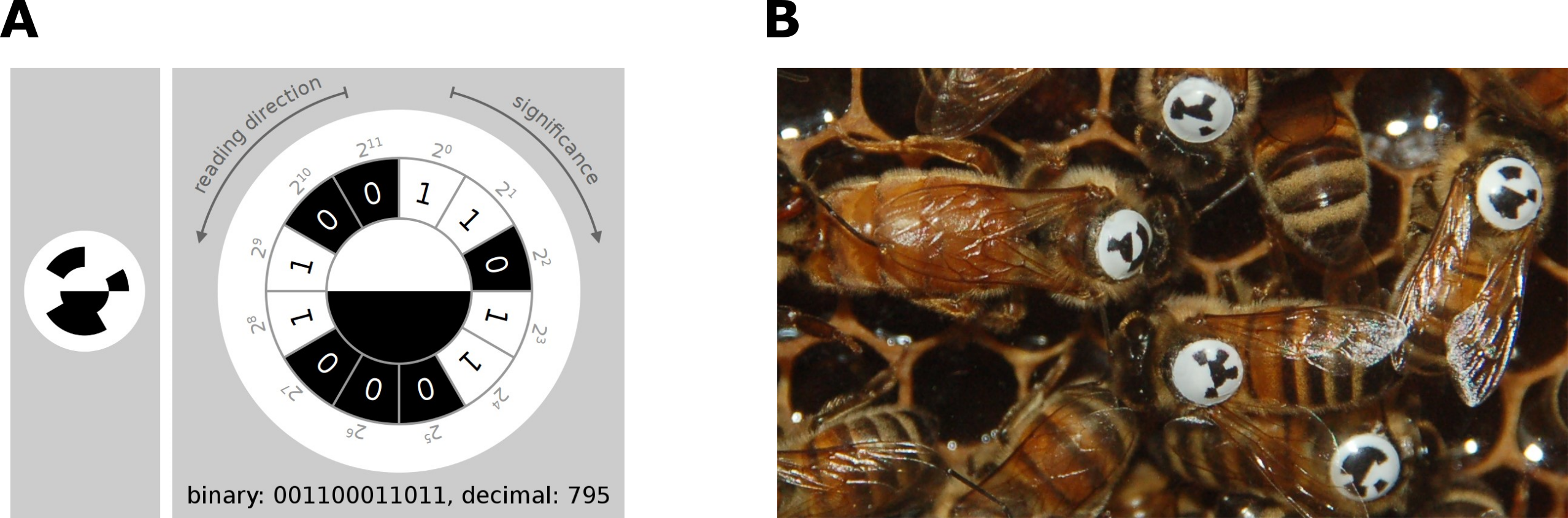}
\end{center}
\caption{(\textbf{A}) The tag-design in the BeesBook project uses 12 coding segments arranged in an arc around two semi-circles that encode the orientation of the bee. The tag is glued onto the thorax such that the white semi-circle is rotated towards the bee's head. Figure adapted from \citep{wario_vazquez_phd_2017}. (\textbf{B}) Several tagged honey bees on a comb. The round and curved tags are designed to endure heavy duty activities such as cell inspections and foraging trips.}\label{fig:tagdesign}
\end{figure}

In contrast to these solutions, we have developed a system called BeesBook that uses much less expensive recording equipment \citep{wario_automatic_2015}. \cref{fig:setup} shows our recording setup, \cref{fig:pipeline} visualizes the processing steps performed after the recording. Our system localizes tags with a recall of 98\% at 99\% precision and decodes 86\% IDs correctly without relying on error correcting codes \citep{wild_automatic_2018}. See  \cref{fig:tagdesign} for the tag design. Linking detections only based on matching IDs would quickly accumulate errors, long-term trajectories would exhibit gaps or jumps between individuals. Following individuals robustly, thus, requires a more elaborate tracking algorithm.

The field of multiple object tracking has produced numerous solutions to various use-cases such as pedestrian and vehicle tracking (for reviews see \cite{cox_review_1993, wu_online_2013, luo_multiple_2014, betke_data_2016}). Animals, especially insects, are harder to distinguish and solutions for tracking multiple animals over long time frames are far less numerous (see \cite{dell_automated_2014} for a review on animal tracking). Since our target subjects may leave the area under observation at any time, the animal's identity cannot be preserved by tracking alone. We require some means of identification for a new detection, whether it be paint marks or number tags on the animals, or identity-preserving descriptors extracted from the detection.

While color codes are infeasible with monochromatic imaging, using image statistics to “fingerprint” sequences of visible animals \citep{perez-escudero_idtracker_2014-1, kuhl_animal_2013, wang_learning_2013} may work even with unstructured paint markers. Merging tracklets after occlusions can then be done by matching fingerprints. However, it remains untested whether these approaches can resolve the numerous ambiguities in long-term observations of many hundreds or thousands of bees that may leave the hive for several hours.

In the following, we describe the features that we used to train machine learning classifiers to link individual detections and short tracklets in a crowded bee hive. We evaluate our results with respect to path and ID correctness. We conclude that long-term tracking can be performed without marker-based error correction codes. Tracking can, thus, be conducted without expensive high-resolution, low-noise camera equipment. Instead, decoding errors in simple markers can be mitigated by the proposed tracking solution, leading to a higher final accuracy of the assigned IDs compared to other marker-based systems that do not employ a tracking step.

\section{Description of Methods}
\label{sec:methods}

\subsection{Problem statement and overview of tracking approach}

The tracking problem is defined as follows: Given a set of detections (timestamp, location, orientation and ID information), find correct correspondences among detections over time (tracks) and assign the correct ID to each track. The ID information of the detections can contain errors. Additionally, correct correspondences between detections of consecutive frames might not exist due to missing detections caused by occluded markers.
In our dataset, the ID information consists of a number in the range of 0 to 4095, represented by 12 bits. Each bit is given as a value between 0.0 and 1.0 which corresponds to the probability that the bit is set.

\begin{figure}[h!]
\begin{center}
    \includegraphics[width=\textwidth]{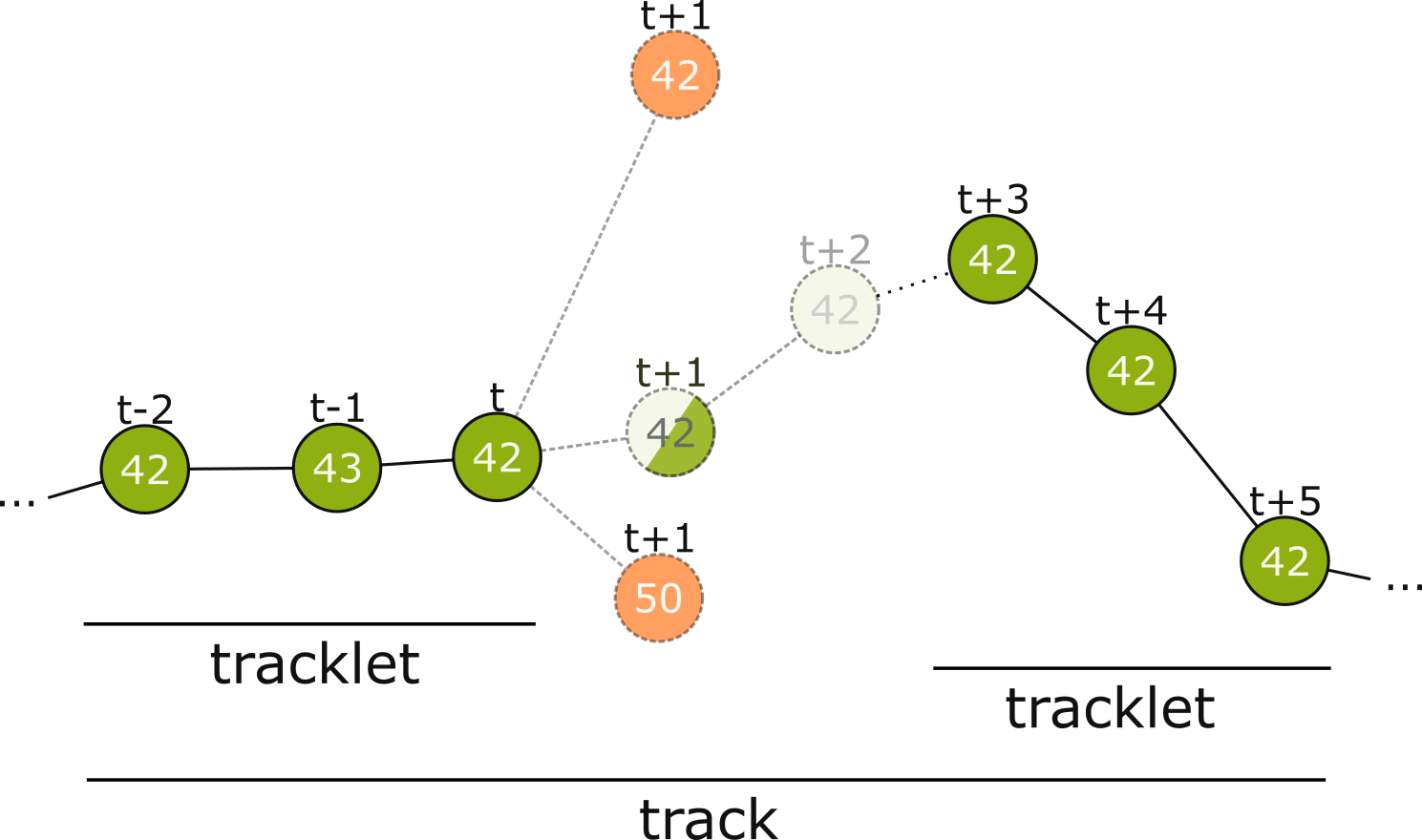} 
\end{center}
\caption{Overview of the tracking process. The first step connects detections from successive frames to tracklets without gaps. At time step t only detections within a certain distance are considered. Even if a candidate has the same ID (top-most candidate with ID 42) it can be disregarded. The correct candidate may be detected with an erroneous ID (see t-1) or may even not be detected at all by the computer vision process. There may be close incorrect candidates that have to be rejected (candidate with ID 43 at t+1). The model assigns a correspondence probability to all the candidates. If none of them receive a sufficient score the tracklet is closed. In  time step t+3 a new detection with ID 42 occurs again and is extended into a second tracklet. In tracking step 2, these tracklets are combined to a larger tracklet or track.
}\label{fig:tracking}
\end{figure}

To solve the described tracking problem, we propose an iterative tracking approach, similar to previous works (for reviews, see \cite{betke_data_2016, luo_multiple_2014}). We use two steps: 1. Consecutive detections are combined into short but reliable tracklets \citep{rosemann_tracking}. 2. These tracklets are connected over longer gaps \citep{BoenischFeatureEngineeringProbabilistic2017}. Previous work employing machine learning mostly scored different distance measures separately to combine them into one thresholded value for the first tracking step \citep{wu_detection_2007, huang_robust_2008, wang_multiple-human_2014, Fasciano2013}. For merging longer tracks, boosting models to predict a ranking between candidate tracklets have been proposed \citep{huang_robust_2008, Fasciano2013}. We use machine learning models in both steps to learn the probability that two detections, or tracklets, correspond. We train the models on a manually labeled dataset of ground truth tracklets. The features that are used to predict correspondence can differ between detection level and tracklet level, so we treat these two stages as separate learning problems.
Both of our tracking steps use the Hungarian algorithm \citep{KuhnHungarianmethodassignment1955} to assign likely matches between detections in subsequent time steps based on the predicted probability of correspondence.
In the following, we describe which features are suitable for each step and how we used various regression models to create accurate trajectories. We also explain how we integrate the ID decodings of the markers along a trajectory to predict the most likely ID for this animal, which can then be used to extract long-term tracks covering the whole lifespan of an individual. See \cref{fig:tracking} for an overview of our approach.

\subsection{Step 1: Linking consecutive detections}

The first tracking step considers detections in successive frames. To reduce the number of candidates, we consider only sufficiently close detections (we use approximately 200 pixels, or 12mm).

From these candidate pairs we extract three features:
\begin{enumerate}
\item Euclidean distance between the first detection and its potential successor.
\item Angular difference of both detections' orientations on the comb plane.
\item Manhattan distance between both detections' ID probabilities.
\end{enumerate}

We use our manually labeled training data to create samples with these features that include both correct and incorrect examples of correspondence. A support vector machine (SVM) with a linear kernel \citep{Cortes1995} is then trained on these samples. We also evaluated the performance of a random forest classifier \citep{HoRandomdecisionforests1995} with comparable results. We use the SVM implemented in the scikit-learn library \citep{pedregosa_scikit-learn_2011}. Their implementation of the probability estimate uses Platt's method \citep{platt_probabilistic_1999}.
This SVM can then be used get the probability of correspondence for pairs of detections that were not included in the training data.
To create short tracks (tracklets), we iterate through the recorded data frame by frame and keep a list of open tracklets. Initially, we have one open tracklet for each detection of the first frame. For every time step, we use the SVM to score all new candidates against the last detection of each open tracklet. The Hungarian algorithm is then used to assign the candidate detections to the open tracklets. Tracklets are closed and not further expanded if their best candidate has a probability lower than $0.5$. Detections that could not be assigned to an existing open tracklet are used to begin a new open tracklet that can be expanded in the next time step.

\subsection{Step 2: Merging tracklets}

The first step yields a set of short tracklets that do not contain gaps and that could be connected with a high confidence. The second tracking step merges these tracklets into longer tracks that can contain gaps of variable duration (for distributions of tracklet and gap length in our data see \cref{sec:results}). Note that a tracklet could consist of a single detection or that its corresponding consecutive tracklet could still begin in the next time step without a gap. To reduce computational complexity we define a maximum gap length of 14 time steps ($\sim$ 4s in our recordings).

Similar to the first tracking step, we use the ground truth dataset to create training samples for a machine learning classifier. We create positive samples (i.e. fragments that should be classified as belonging together) by splitting each manually labeled track once at each time step. Negative samples are generated from each pair of tracks with different IDs which overlapped in time with a maximum gap size of 14. These are also split at all possible time steps. To include both more positive samples and more short track fragments in the training data, we additionally use every correct sub-track of length 3 or less and again split it at all possible locations.
This way we generated 1.021.848 training pairs, 7.4\% of which were positive samples.

In preliminary tests, we found that for the given task of finding correct correspondences between tracklets, a random forest classifier performed best among a selection of classifiers available in scikit-learn \citep{BoenischFeatureEngineeringProbabilistic2017}.  

Tracklets with two or more detections allow for more complex and discriminative features compared to those used in the first step. For example, matching tracklets separated by longer gaps may require features that reflect a long-term trend (e.g. the direction of motion).

We implemented 31 different features extractable from tracklet pairs.
We then used four different feature selection methods from the scikit-learn library to find the features with the highest predictive power. This evaluation was done by splitting the training data further into a smaller training set and validation set. The methods used were Select-K-Best, Recursive Feature Elimination, Recursive Feature Elimination with Cross-Validation and the Random Forest Feature Importance for all possible feature subset sizes as provided by scikit-learn \citep{scikit-learn}. In all these methods, the same four features (number 1 to 4 in the listing below) performed best according to the ROC AUC score \citep{spackman_signal_1989} that proved to be a suitable metric to measure tracking results. Therefore, we chose them as an initial subset.

\begin{figure}[h!]
\begin{center}
    \includegraphics[width=\textwidth]{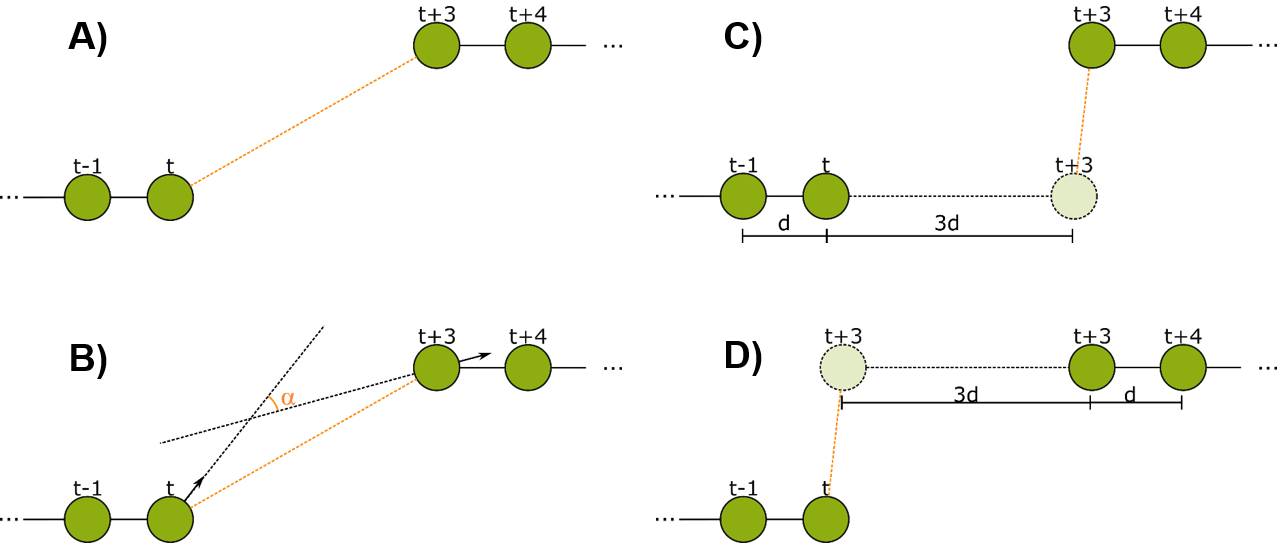} 
\end{center}
\caption{The spatial features used in  the second tracking step. A) Euclidean distance between the last detection of tracklet 1 and the first detection of tracklet 2. B) Forward error: Euclidean distance of the extrapolation of the last movement vector in tracklet 1 to the first detection in tracklet 2. C) Angular difference between the tag orientations of the last detection in tracklet 1 and the first detection in tracklet 2. D) Backward error: Euclidean distance between the reverse extrapolation of the first movement vector of tracklet 2 to the last detection of tracklet 1.}\label{fig:features}
\end{figure}

We then tried to improve the feature subset manually according to more tracking-specific metrics. The metrics we used were the number of tracks in the ground truth validation set that were reconstructed entirely and correctly, and the number of insertions and deletes in the tracks (for further explanation of the metrics see \cref{sec:results}).  We added the features that lead to the highest improvements in these metrics on our validation set. This way, we first added feature 5 and then 6. After adding feature 6, the expansion of the subset with any other feature only lead to a performance decrease in form of more insertions and less complete tracks. We therefore kept the following six features. Visualizations of features 2 to 5 can be found in \cref{fig:features}.

\begin{enumerate}
\item Manhattan distance of both tracklets' bitwise averaged IDs.
\item Euclidean distance of last detection of tracklet 1 to first detection of tracklet 2.
\item Forward error: Euclidean distance of linear extrapolation of last motion in first tracklet to first detection in second tracklet.
\item Backward error: Euclidean distance of linear extrapolation of first motion in second tracklet to last detection in first tracklet.
\item Angular difference of tag orientation between the last detection of the first tracklet and the first detection of the second tracklet.
\item Difference of confidence: all IDs in both tracklets are averaged with a bitwise median, we select the bit that is closest to $0.5$ for each tracklet, calculate the absolute difference to $0.5$ (the confidence) and compute the absolute difference of these two confidences.
\end{enumerate}

\subsubsection{Track ID assignment}

After the second tracking step, we determine the ID of the tracked bee by calculating the median of the bitwise ID probabilities of all detections in the track. The final ID is then determined by binarizing the resulting probabilities for each bit with probability threshold 0.5.

\subsubsection{Parallelization}

Tracks with a length of several minutes already display a very accurate ID decoding (see Section \ref{sec:results}). To calculate longer tracks of up to several days and weeks, we execute the tracking step 1 and step 2 for intervals of one hour and then merge the results to longer tracks based on the assigned ID. This allows us to effectively parallelize the tracking calculation and track the entire season of ten weeks of data in less than a week on a small cluster with less than 100 CPU cores.

\section{Results and evaluation}
\label{sec:results}

We marked an entire colony of 1953 bees in a two days session and continuously added marked young bees that were bred in an incubation chamber. In total, 2775 bees were marked. The BeesBook system was used to record 10 weeks of continuous image data (3 Hz sample rate) of a one-frame observation hive. The image recordings were stored and processed after the recording season. The computer vision pipeline was executed on a Cray XC30 supercomputer. In total, 3,614,742,669 detections were extracted from 67,972,617 single frames, corresponding to 16,993,154 snapshots of the four cameras. Please note that the data could also be processed in real-time using consumer hardware \citep{wild_automatic_2018}.

Two ground truth datasets for the training and evaluation of our method were created manually. A custom program was used to mark the positions of an animal and to define its ID \citep{mischek_probabilistisches_2016}. Details on each dataset can be found in Table \ref{tab:datasets}. To avoid overfitting to specific colony states, the datasets were chosen to contain both high activity (around noon) and low activity (in the early morning hours) periods, different cameras and, therefore, different comb areas. Dataset \textit{2015.1} was used to train and validate classifiers and dataset \textit{2015.2} was used to test their performance.

\begin{table}[]
\centering
\begin{tabular}{c|c|c}
Dataset & 2015.1 & 2015.2\tabularnewline
\hline
\hline
Date & 18.09.2015 & 22.09.2015\tabularnewline
\hline
Times & 11:36; 04:51 & 13:36\tabularnewline
\hline
Frames & 201 (3 fps) & 200 (3 fps)\tabularnewline
\hline
Detections & 18085 & 10945\tabularnewline
\hline
False positives & 222 (1.23\%) & 82 (0.75\%)\tabularnewline
\hline
Individuals & 144 & 98\tabularnewline

\end{tabular}
\caption{Dataset \textit{2015.1} was used for training and dataset \textit{2015.2} for testing. The number of detections is the number of tags localized and decoded by the deep learning approach over all frames in the dataset. The number of false positives shows how many times the deep learning pipeline detects a detection when there is none. The number of individuals indicates how many different bees are present in the dataset.}\label{tab:datasets}
\end{table}

Dataset \textit{2015.1} contains 18085 detections from which we extracted 36045 sample pairs (i.e. all pairs with a distance of less than 200 pixels in consecutive frames). These samples were used to train the SVM which is used to link consecutive detections together (tracking step 1). Hyperparameters were determined manually using cross-validation on this dataset. The final model was evaluated on dataset \textit{2015.2}.

Tracklets for the training and evaluation of a random forest classifier (tracking step 2) were extracted from datasets \textit{2015.1} respectively \textit{2015.2} (see Section \ref{sec:methods} for details). Hyperparameters were optimized with hyperopt-sklearn \citep{komer_hyperopt-sklearn:_2014} on dataset \textit{2015.1} and the optimized model was then tested on dataset \textit{2015.2}.

To validate the success of the tracking, we analyzed its impact on several metrics in the tracks, namely:
\begin{enumerate}
\item ID Improvement
\item Proportion of complete tracks
\item Correctness of resulting tracklets
\item Length of resulting tracklets
\end{enumerate}
To be able to evaluate the improvement through the presented iterative tracking approach, we compare the results of the two tracking steps to the naive approach of linking the original detections over time based on their initial decoded ID only, in the following referred to as ``baseline''. For an overview on the improvements achieved by the different tracking steps see \cref{tab:2}.

\begin{table}[]
\centering
\begin{tabular}{l|c|c|c|c}
 & baseline  & after step 1 & after step 2  & perfect tracking \\
 \hline
 \hline
 incorrect detection IDs & 13.3\%  & 3.9\% & 1.9\% & 0.6\% \\
 \hline
 incorrect track IDs & 63.5 & 27.2\% & 18.2\% & 8.2\% \\
 \hline
 complete tracks & 10.2\% & 26.5\% & 70.4\% & 77.6\% \\
 \hline
 \begin{tabular}{@{}l@{}}detections missing from\\ their track (deletions)\end{tabular} & 32.2\% & 1.38\% & 2.37\% & 0\% \\
 \hline
 \begin{tabular}{@{}l@{}}tracks with at least\\ one deletion\end{tabular} & 94.6\% & 26.7\% & 18.25\% & 0\%  
\end{tabular}
\caption{Different metrics were used to compare the two tracking steps to both a naive baseline based on the detection IDs and to manually created tracks without errors (perfect tracking). In all cases, the baseline performs worst and the two tracking steps successively improve the performance.}\label{tab:2}
\end{table}

\paragraph*{ID improvement}
An important goal of the tracking is to correct IDs of detections which could not be decoded correctly by the computer vision system. Without the tracking algorithm described above, all further behavioral analyses would have to consider this substantial proportion of erroneous decodings. In our dataset, 13.3\% of all detections have an incorrectly decoded ID \citep{wild_automatic_2018}.

\begin{figure}[h!]
\begin{center}
    \includegraphics[width=10cm]{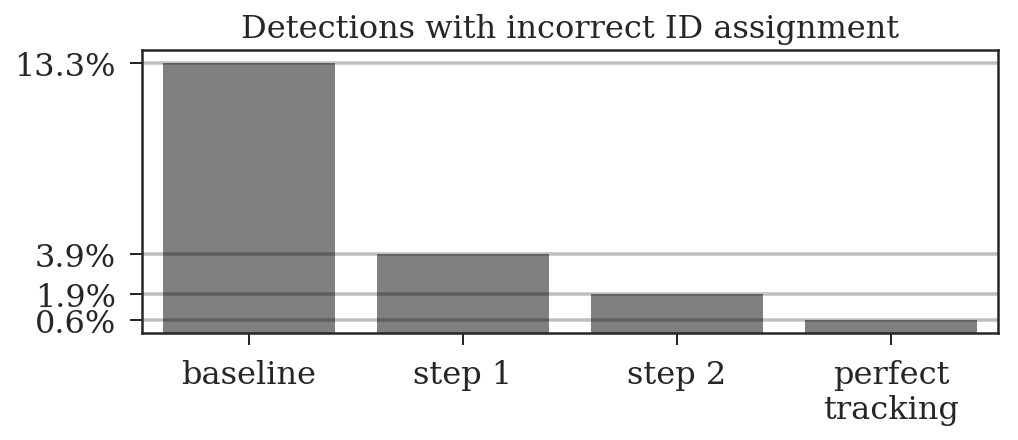}
\end{center}
\caption{Around 13\% of the raw detections are incorrectly decoded. The first tracking step already reduces this error to around 4\% and the second step further reduces it to around 2\%. Even a perfect tracking (defined by the human ground truth) would still result in 0.6\% incorrect IDs when using the proposed ID assignment method.}\label{fig:detectionincorrect}
\end{figure}

In the ground truth dataset we manually assigned detections that correspond to the same animal to one trajectory. The ground truth data can therefore be considered as the ``perfect tracking''. Even on these perfect tracks the median ID assignment algorithm described above provides incorrect IDs for 0.6\% of all detections, due to partial occlusions, motion blur and image noise. This represents the lower error bound for the tracking system.
As shown in \cref{fig:detectionincorrect}, the first tracking step reduces the fraction of incorrect IDs from 13.3\% to 3.9\% of all detections. The second step further improves this result to only 1.9\% incorrect IDs. 

\begin{figure}[h!]
\begin{center}
    \includegraphics[width=10cm]{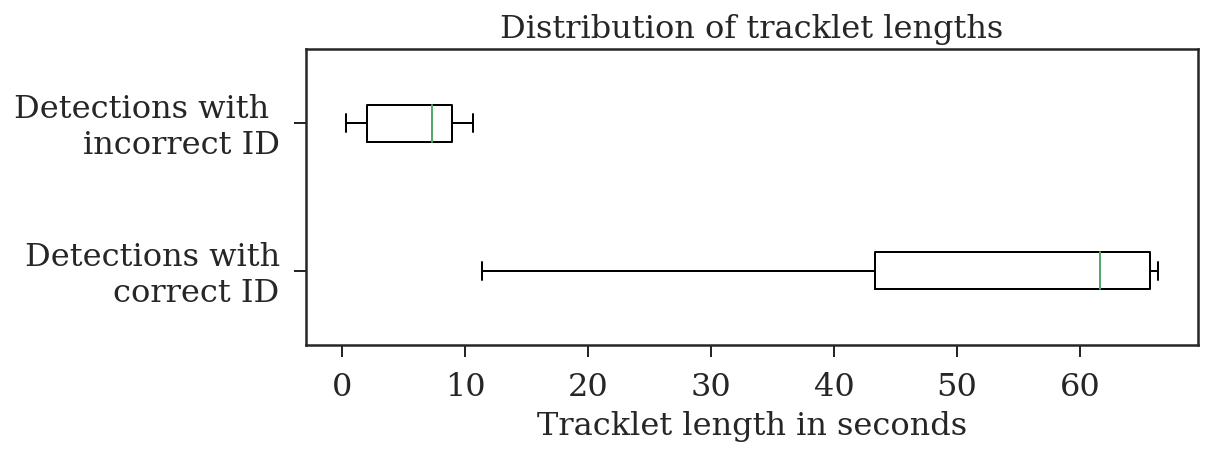}
\end{center}
\caption{Evaluation of the tracklet lengths of incorrectly assigned detection IDs after the second tracking step reveals that all errors in the test dataset \emph{2015.2} happen in very short tracklets. Note that this dataset covers a duration of around one minute.}\label{fig:trackleterrorlength}
\end{figure}

\begin{figure}[h!]
\begin{center}
    \includegraphics[width=10cm]{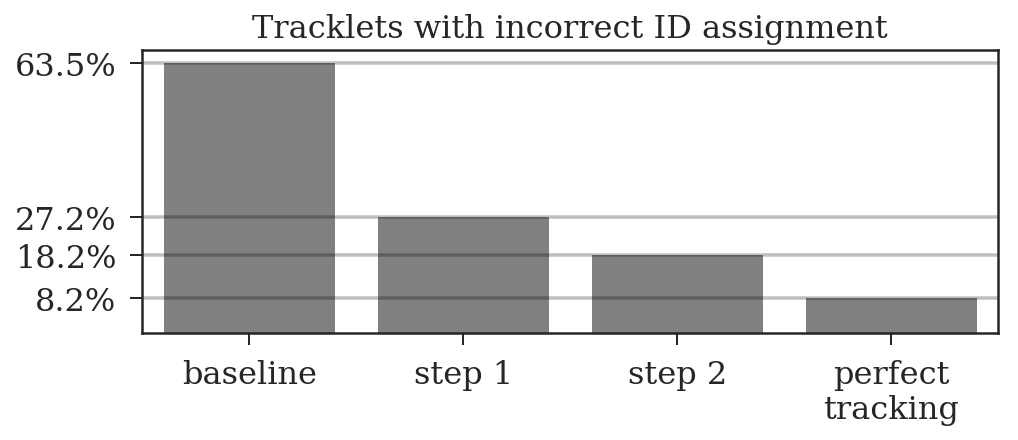}
\end{center}
\caption{A naive tracking approach using only the detection IDs would result in around 64\% of all tracks being assigned an incorrect ID. Our two-step tracking approach reduces this to around 27\% and 18\% respectively. Due to the short length of most incorrect tracklets, these 18.2\% account for only 1.9\% of the detections. Using our ID assignment method without any tracking errors would reduce the error to 8.2\%.}\label{fig:trackletsincorrect}
\end{figure}

Most errors occur in short tracklets (see Figure \ref{fig:trackleterrorlength}). Therefore, the 1.9\% erroneous ID assignments correspond to 18.2\% of the resulting tracklets being assigned an incorrect median ID. This is an improvement over the naive baseline and the first tracking step with 63.5\% and 27.2\% respectively. A perfect tracking could reduce this to 8.2\% (see \cref{fig:trackletsincorrect}).

\paragraph*{Proportion of complete tracks}

\begin{figure}[h!]
\begin{center}
    \includegraphics[width=10cm]{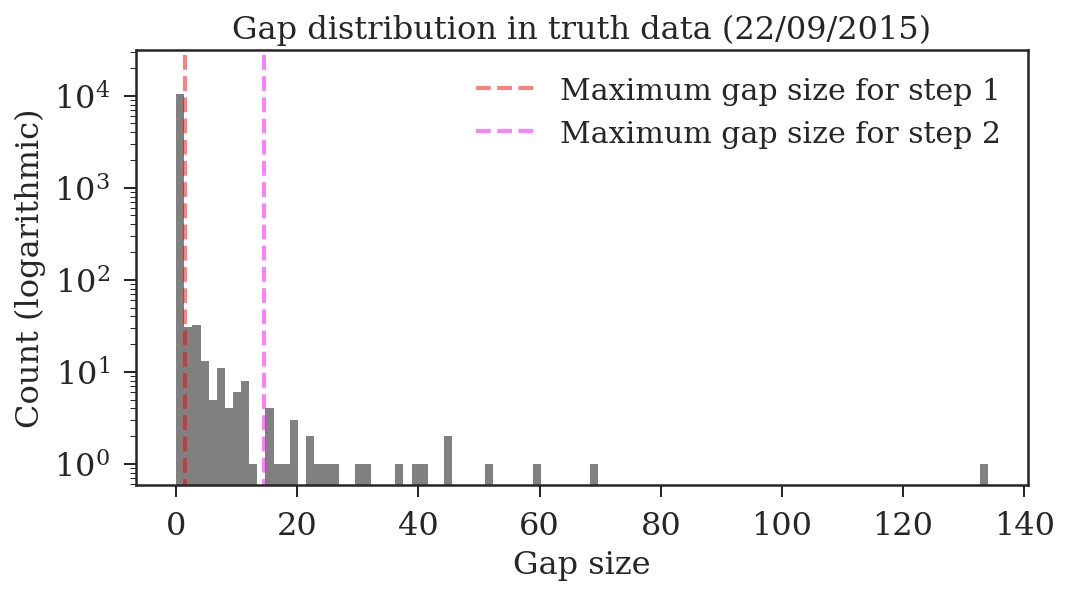}
\end{center}
\caption{Distribution of the gap sizes in the ground truth dataset \textit{2015.2}. Most corresponding detections (i.e. 97.9\%) have no gaps and can be therefore be matched by the first tracking step. The resulting tracklets are then merged in the second step. The maximum gap size of 14 covers 99.76\% of the gaps.}\label{fig:gapsize}
\end{figure}

\begin{figure}[h!]
\begin{center}
    \includegraphics[width=10cm]{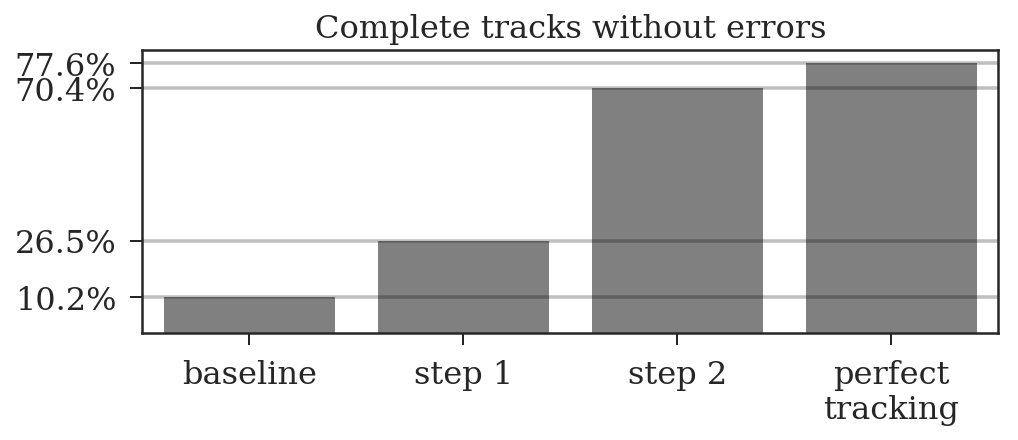}
\end{center}
\caption{A \textit{complete track} perfectly reconstructs a track in our ground truth data without any missing or incorrect detections. Even a perfect tracking that is limited to a maximum gap size of 14 frames could only reconstruct around 78\% of these tracks. The naive baseline based only on the detection IDs would assemble 10\% without errors while our two tracking steps achieve 26.5\% and 70.4\% respectively.}\label{fig:trackingquality}
\end{figure}

Almost all gaps between detections in our ground truth tracks are no longer than 14 frames (99.76\%, see \cref{fig:gapsize}). Even though large gaps between detections are rare, long tracks are likely to contain at least one such gap: Only around one third (34.7\%) of the ground truth tracks contain no gaps and 77.6\% contain only gaps shorter than 14 frames.
As displayed in \cref{fig:trackingquality}, the baseline tracking finds only 10.2\% complete tracks without errors (i.e. 30\% of all tracks with no gaps). Step 1 is able to correctly assemble 26.5\% complete tracks (i.e. around 76.5\% of all tracks containing no gaps). Step 2 correctly assembles 70.4\% complete tracks (about 90.4\% of all tracks with a maximum gap size of less than 14 frames).

\paragraph*{Correctness of resulting tracklets}

To characterize the type of errors in our tracking results, we define a number of additional metrics.
We counted detections that were incorrectly introduced into a track as \textit{insertions}. Both tracking steps and the baseline inserted only one incorrect detection into another tracklet. Thus less than 1\% of both detections and tracklets were affected.

We counted detections that were missing from a tracklet (and were replaced by a gap) as \textit{deletions}. In the baseline, 32.2\% of all detections were missing from their corresponding track (94.6\% of all tracks had at least one deletion). After the first step, 1.38\% of detections were missing from their track, affecting 26.7\% of all tracks. After the second step, 2.37\% of all detections and 18.25\% of all tracks were still affected.

We also evaluated whether incorrect detections were contained in a track in situations where the correct detection would have been available (instead of a gap) as \textit{mismatches}, but no resulting tracks contained such mismatches.

\paragraph*{Length of resulting tracklets}

\begin{figure}[h!]
\begin{center}
    \includegraphics[width=10cm]{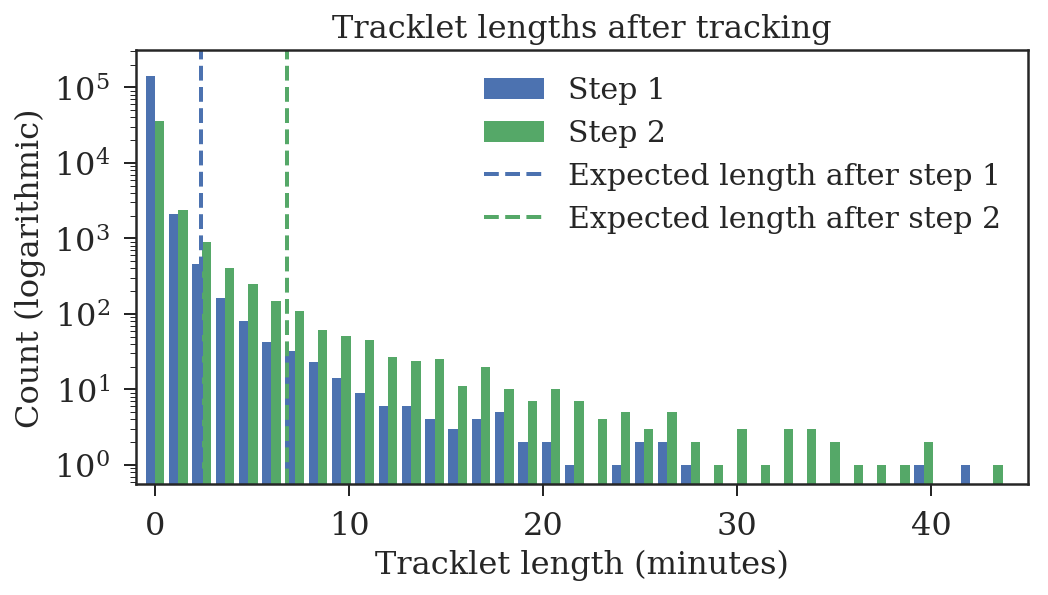}
\end{center}
\caption{Track lengths after tracking one hour of video data at three frames per second. The expected length of a track is 2:23 minutes after the first step and 6:48 minutes after the second step.}\label{fig:tracklength}
\end{figure}

The ground truth datasets contain only short tracks with a maximum length of one minute. To evaluate the average length of the tracks, we also tracked one hour of data for which no ground truth data was available. The first tracking step yields shorter fragments with an expected length of 2:23 minutes, the second tracking step merges these fragments to tracklets with an expected duration of 6:48 minutes (refer to Figure ~\ref{fig:tracklength} for tracklet length distributions).

\section{Discussion}

We have presented a multi-step tracking algorithm for fragmentary and partially erroneous detections of honey bee markers. We have applied the proposed algorithm to produce long-term trajectories of all honey bees in a colony of approximately 2000 animals. Our dataset comprises 71 days of continuous positional data at a recording rate of 3 Hz. The presented dataset is by far the most detailed reflection of individual activities of the members of a honey bee colony. The dataset covers the entire lifespan of many hundreds of animals from the day they emerge from their brood cell until the day they die. Honey bees rely on a flexible but generally age-dependent division of labor. Hence, our dataset reflects all essential aspects of a self-sustaining colony, from an egg-laying queen and brood rearing young workers, to food collection, and colony defense. We have released a three days sample dataset for the interested reader \citep{boenisch_beesbook_2017}. Our implementation of the proposed tracking algorithm is available online\footnote{\url{https://github.com/BioroboticsLab/bb_tracking}}.

The tracking framework presented in the previous sections is an essential part of the BeesBook system. It provides a computationally efficient approach to determine the correct IDs for more than 98\% of the individuals in the honey bee hive without using extra bits for error correction.

Although it is possible to use error correction with 12 bit markers, this would reduce the number of coding bits and therefore the number of observable animals. While others chose to increase the number of bits on the marker, we solved the problem in the tracking stage. With the proposed system, we were able to reduce hardware costs for cameras and storage. When applied to the raw output of the image decoding step, the accuracy of other systems that use error-correction (for example \cite{Mersch2013}) may even be improved further.

Our system provides highly accurate movement paths of bees. Given a long-term observation of several weeks, these paths, however, can still be considered short fragments. Since the IDs of these tracklets are very accurate, they can now be linked by matching IDs only.

Still, some aspects of the system can be improved. To train our classifiers, we need a sufficiently large, manually labeled dataset. \cite{rice_efficient_2015} proposed a method to create a similar dataset interactively, reducing the required manual work. Also, the circular coding scheme of our markers causes some bit configurations to appear similar under certain object poses. This knowledge could be integrated into our ID determination algorithm. The IDs along a trajectory might not provide an equal amount of information. Some might be recorded under fast motion and are therefore less reliable. Other detections could have been recorded from a still bee whose tag was partially occluded. Considering similar readings as less informative might improve the ID accuracy of our method. Still, with the proposed method there are only 1.9\% detections incorrectly decoded, mostly in very short tracklets.

The resulting trajectories can now be used for further analyses of individual honey bee behavior or interactions in the social network. In addition to the three day dataset published alongside this paper, we plan to publish two more datasets covering more than 60 days of recordings, each. With this data we can investigate how bees acquire information in the colony and how that experience modulates future behavior and interactions. We hope that through this work we can interest researchers to join the collective effort of investigating the individual and collective intelligence of the honey bee, a model organism that bears a vast number of fascinating research questions.

\section*{Conflict of Interest Statement}

The authors declare that the research was conducted in the absence of any commercial or financial relationships that could be construed as a potential conflict of interest.

\section*{Funding}
FW received funding from the German Academic Exchange Service (DAAD). DD received funding from the Andrea von Braun Foundation. This work was in part funded by the Klaus Tschira Foundation. We also acknowledge the support by the Open Access Publication Initiative of the Freie Universität Berlin.

\section*{Acknowledgments}
We are indebted to the help of Jakob Mischek for his preliminary work and his help with creating the ground truth data.

\bibliography{library}

\bibliographystyle{plainnat}

\end{document}